\documentclass[10pt,twocolumn,letterpaper]{article}

\usepackage{comment}
\usepackage{wacv}              

\definecolor{wacvblue}{rgb}{0.21,0.49,0.74}
\usepackage[pagebackref,breaklinks,colorlinks,allcolors=wacvblue]{hyperref}

\title{CubeletWorld: A New Abstraction for Scalable 3D Modeling}

\author{Azlaan Mustafa Samad\\
L3S Research Center\\
Leibniz University Hannover, Germany\\
{\tt\small azlaan.samad@l3s.de}
\and  
Hoang H. Nguyen\\
University of Tennessee at Chattanooga\\
Tennessee, USA\\
{\tt\small huuhoang-nguyen@utc.edu}
\and
Lukas Berg\\
L3S Research Center\\
Leibniz University Hannover, Germany\\
{\tt\small berg@tnt.uni-hannover.de}
\and
Henrik Müller\\
L3S Research Center\\
Leibniz University Hannover, Germany\\
{\tt\small hmueller@l3s.de }
\and
Yuan Xue\\
L3S Research Center\\
Leibniz University Hannover, Germany\\
{\tt\small xue@l3s.de}
\and
Daniel Kudenko\\
L3S Research Center\\
Leibniz University Hannover, Germany\\
{\tt\small kudenko@l3s.de}
\and
Zahra Ahmadi\\
PLRI Medical Informatics Institute \\
Hannover Medical School, Germany\\
{\tt\small ahmadi.zahra@mh-hannover.de}
}

\begin{document}
\maketitle
\begin{abstract}
Modern cities produce vast streams of heterogeneous data, from infrastructure maps to mobility logs and satellite imagery. However, integrating these sources into coherent spatial models for planning and prediction remains a major challenge. Existing agent-centric methods often rely on direct environmental sensing, limiting scalability and raising privacy concerns. 
This paper introduces \textit{CubeletWorld}, a novel framework for representing and analyzing urban environments through a discretized 3D grid of spatial units called cubelets. This abstraction enables privacy-preserving modeling by embedding diverse data signals, such as infrastructure, movement, or environmental indicators, into localized cubelet states. CubeletWorld supports downstream tasks such as planning, navigation, and occupancy prediction without requiring agent-driven sensing.
To evaluate this paradigm, we propose the CubeletWorld State Prediction task, which involves predicting the cubelet state using a realistic dataset containing various urban elements like streets and buildings through this discretized representation. We explore a range of modified core models suitable for our setting and analyze challenges posed by increasing spatial granularity, specifically the issue of sparsity in representation and scalability of baselines. In contrast to existing 3D occupancy prediction models, our cubelet-centric approach focuses on inferring state at the spatial unit level, enabling greater generalizability across regions and improved privacy compliance. Our results demonstrate that CubeletWorld offers a flexible and extensible framework for learning from complex urban data, and it opens up new possibilities for scalable simulation and decision support in domains such as socio-demographic modeling, environmental monitoring, and emergency response. The code and datasets can be downloaded from \href{https://drive.google.com/file/d/1KwptXrh_fbq7JGCJFP1L-f2Y3Vfe2UZC/view?usp=sharing}{here.}
\end{abstract}
\section{Introduction}\label{sec1}

\begin{figure}[tb]
\centering
\includegraphics[width=0.95\linewidth, keepaspectratio]{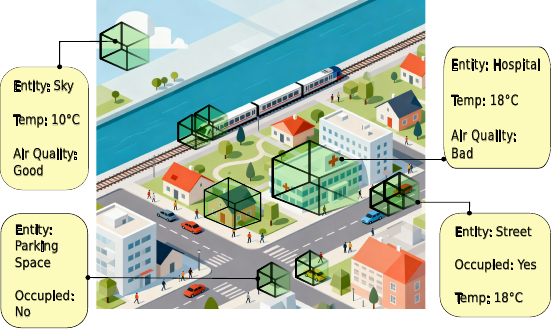}
\caption{An example of CubeletWorld with cubelets spanning across different entities across the city. Different cubelets contain different types of information about the volumetric space, such as entity type, occupancy, temperature, and air quality index. Different cubelets can be of different dimensions. }
\label{multimodal}
\end{figure}

Understanding and forecasting the dynamic state of complex urban environments is critical to applications in planning, occupancy prediction, and emergency response. Existing approaches often rely on 2D maps or agent-centric sensor data, which capture only fragments of the full three-dimensional reality and lack the structure and scalability required for consistent, multi-purpose inference across time and space. 
We introduce \textit{CubeletWorld}, a discretized 3D grid-based abstraction of urban space (Figure~\ref{multimodal}), where the environment is partitioned into uniform volumetric units, cubelets, each encoding relevant attributes such as occupancy, temperature, pollution, or sensor-derived signals. CubeletWorld regularizes fragmented sensor data into cubelets, enabling inference even in regions without direct observations. This supports predictions of both occupancy and broader environmental dynamics. Its multi-resolution capability allows cubelets to represent either small volumes (e.g., 3×3×3 m) or entire districts. This supports both fine-grained local reasoning (e.g., congestion on a single street) and coarse-grained global analysis (e.g., city-wide traffic or pollution). Importantly, cubelets preserve privacy by abstracting away from individual trajectories.

Existing data sources, such as cameras, air quality monitors, or mobile devices, offer only partial, pointwise measurements. By discretizing space into cubelets, these signals can be aggregated and interpolated, enabling inference in regions with sparse or missing data. For instance:
\begin{itemize} [leftmargin=*,topsep=0pt]
    \item[-] Pollution control: sensors may measure air quality at specific intersections; CubeletWorld fills gaps by modeling spatial correlations across cubelets, helping planners identify high-risk areas.
    \item[-] Pandemic spread: instead of tracking individuals, occupancy cubelets reveal evolving population density patterns, useful for simulating exposure risks.
    \item[-] Socioeconomic studies: cubelets can encode census data such as income levels, enabling spatial correlation with mobility or infrastructure usage.
\end{itemize}
This abstraction grounds diverse data streams in a common volumetric framework, making predictions interpretable and actionable. 

CubeletWorld complements \textit{vision-based autonomous driving} research: whereas prior work predicts occupancy around a single agent from multi-view imagery or LiDAR streams \cite{zuo2023pointocc, li2022bevformer, huang2023tri}, we propose to aggregate heterogeneous vision and sensor data into a global 3D grid representation. This cubelet-centric view decouples inference from individual trajectories while retaining compatibility with vision-derived inputs, addressing both scalability and privacy concerns. Similarly, trajectory forecasting approaches \cite{Pellegrini2009YoullNW, Lerner2007CrowdsBE, robicquet2016learning, alahi2016social, donahue2015long} depend on fine-grained identity-linked motion data. CubeletWorld abstracts away from individuals, enabling collective prediction without storing sensitive personal information. Comparable limitations appear in UAV vision research, where works such as Rozantsev et al.~\cite{rozantsev2016detecting}, Coluccia et al.~\cite{coluccia2021drone}, and Kellenberger et al.~\cite{kellenberger2019half} propose robust detection or monitoring methods, but either do not release or only provide restricted-access datasets. These parallels highlight that many vision-based systems, including ours, must operate under assumptions of local-only or non-public data.

We formalize a new family of tasks under \textit{CubeletWorld State Prediction}, where the goal is to forecast future states of cubelets (e.g., occupancy, temperature, or pollution) from historical data. In this paper, we focus on \textit{CubeletWorld occupancy prediction}, predicting which cubelets are likely to be occupied in the future. To enable systematic study, we construct the \textit{CubeletWorld Boids dataset}, a synthetic yet realistic urban landscape populated by Boid agents. This dataset provides controlled conditions for testing cubelet-based forecasting methods before real-world deployment. To systematically study cubelet-centric reasoning in a controlled setting, we introduce the CubeletWorld Boids dataset. While real-world cubelet-level datasets are not yet available, this synthetic yet realistic proxy enables rigorous exploration of key challenges such as sparsity and scalability, and lays the groundwork for future real-world deployments. The dataset illustrates spatiotemporal reasoning at multiple resolutions, where the goal is to predict which cubelets are occupied by Boids over time. 

We benchmark forecasting models on CubeletWorld and identify fundamental challenges, including handling sparsity, scaling to high-resolution grids, and reasoning across multiple spatial resolutions. Our contributions are:
\begin{itemize}[leftmargin=*]
    \item Representation: We propose \textit{CubeletWorld}, a novel unified 3D representation for storing and reasoning about physical environments using discretized volumetric data. 
    \item Task family: We define and formalize the \textit{CubeletWorld State Prediction}, a set of forecasting tasks applicable across urban, environmental, and privacy-aware modeling domains.
    \item Dataset and benchmark: We introduce \textit{CubeletWorld Boids dataset} and provide initial benchmarks, exposing fundamental challenges in cubelet-centric learning. 
\end{itemize}

\section{Problem Definition}\label{motivationandproblemdefinition}

CubeletWorld enables the storage of diverse data types within a discretized volumetric space, providing a foundation for predictive modeling of the cubelet's future state. In this work, we introduce a novel state prediction task in the CubeletWorld: \textit{CubeletWorld occupancy prediction}. We propose a generalizable framework for this prediction task, which can be readily extended to other prediction objectives based on the types of data stored in the cubelets.
The task, \textit{CubeletWorld occupancy prediction}, aims to forecast the occupancy status of each cubelet using historical information of the environment. This task falls under a broader class of state prediction problems in CubeletWorld, which we describe below. 

\begin{figure}[tb]
\centering
\includegraphics[width=0.9\linewidth]{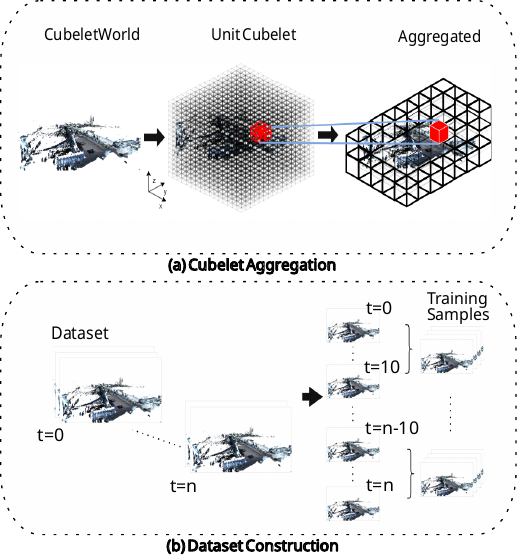}
\caption{CubeletWorld Dataset Preprocessing: (a) The 3D CubeletWorld $\mathcal{E}$ is discretized into total $n$ cubelets each of unit length, this is a matrix  $\mathcal{M}_t \in R^{n_1\times n_2\times n_3}$. In this example, we aggregate eight unit cubelets into a bigger single cubelet. The aggregation of cubelets depends on the choice of resolution required for the use case. In this case, several unit cubelets are aggregated into a single bigger cubelet. The occupancy of the aggregated cubelet depends on the occupancy of the unit cubelet; that is, if any of the unit cubelets is occupied, then the aggregated cubelet is labeled as occupied. (b) The time-series Cubelet Boids dataset is converted to training samples by considering 10 time steps in history and combining them into a single training sample to be used as input to the model.} \label{fig:cubeletdataset}
\end{figure}

We discretize the entire environment $\mathcal{E}$ into $\mathcal{\textit{n}}$ cubeletes. Each cubelet $ \mathcal{C}$ holds information about its state, such as occupancy and temperature, at a given time step $t$, represented either as labels (e.g., binary: $\{0, 1\}$ or multi-class labels $\{0, 1, 2\}$) or continuous values (e.g., sensor or temperature readings). The discretized environment at time step $t$ is represented as a matrix $\mathcal{M}_t \in R^{n_1\times n_2\times n_3}$, where $n_1$, $n_2$ and $n_3$ indicate the number of cubelets in $x$, $y$, and $z$ axes, respectively as shown in Figure \ref{fig:cubeletdataset}. 
To make the state prediction, the model takes into account a sequence of past matrices over previous $t_1$ timesteps, forming a 4D input tensor $\mathcal{M}_t \in R^{t_1\times n_1\times n_2\times n_3}$. The goal of the task is to predict the future state (label or value) for each cubelet for one or more upcoming time steps. 
The number and size of the cubelets can be adjusted based on the desired level of granularity, and are equal to $ n_1\times n_2\times n_3$. A lower resolution (or granularity) implies fewer, larger cubelets, each encompassing a broader spatial area, while a higher resolution allows for fine-grained modeling with a near one-to-one correspondence between $(x, y, z)$ coordinates and cubelets. At the coarsest level, the entire CubeletWorld may be represented by a single cubelet; at the finest level, each coordinate maps directly to its own cubelet.

The high-level objective of this work is to introduce and validate a novel cubelet-centric paradigm for predictive modeling in a discretized 3D environment. Unlike traditional agent-centric approaches, our framework provides a unified spatial perspective that enhances both interpretability and data privacy. Our specific goals are to:
\begin{itemize}[leftmargin=*,topsep=0pt]
    \item Introduce a novel representation of volumetric space in a 3D discretized grid world, which we term \textit{CubeletWorld}.
    \item Demonstrate its utility across various domains, including occupancy prediction, weather forecasting, autonomous navigation, planning, hazardous scenario assessment, and socio-demographic analysis.
    \item Highlight the core advantages of this approach, namely \textit{unified view of the environment}, \textit{multi-modal data} storage capacity, and flexibility in operating at different \textit{levels of granularity}.
\end{itemize}

\section{Datasets}\label{datasets}
In order to make predictions on the two tasks, we first created the \textit{Cubelet Boids dataset}, corresponding to the cubelet occupancy prediction. 

\textit{Cubelet Boids dataset} simulates the movement of agents known as \textit{boids} (short for Bird + oids) within a realistic 3D cubelet environment. The boids simulation mimics flocking behavior observed in birds, using an individual-based model where each boid's (alternatively, bird's) position, velocity, and orientation are governed by three core behavioral rules:
\begin{itemize}[leftmargin=*]
    \item \textit{Separation}: Boids avoid crowding by maintaining a minimum distance from neighbors to avoid collision.
    \item \textit{Alignment}: Boids align their movement direction with nearby flockmates. 
    \item \textit{Cohesion}: Boids steer towards the average position of neighboring boids. 
\end{itemize}
The boid's neighbourhood plays an important role in determining the behavior of each individual boid, and is defined by a spatial arc characterized by a distance threshold (measured from the center of the boid) and a directional angle aligned with its velocity vector. Boids located within this arc influence each other's trajectories.

The \textit{CubeletWorld Boids dataset} consists of two components: the \textit{Terrain dataset} and the \textit{Boids coordinate dataset}. The Terrain dataset comprises fixed entities such as streets, buildings, and vegetation, while the boids coordinate dataset captures the dynamic movement of the boids through the environment. To create the simulated environment, we first set the positions of fixed entities, then include the simulated boid's movement in the CubeletWorld, following the above laws and avoiding crashing with fixed entities. More details of each component are described below.

\subsection{Terrain dataset} The Terrain dataset encodes the static features of CubeletWorld, where each entity has a fixed position $(x, y, z)$ throughout the simulation. The entities include streets, buildings, lamp-posts, and trees, collectively referred to as \textit{terrain}. As shown in Figure \ref{CWspreadOut}, the environment includes tall buildings, green tree cover, numerous small scattered structures, and a horizontally aligned street.
Each terrain coordinate is annotated with \textit{RGB} values (indicating color) and \textit{Reflectance} values (indicating surface light reflectivity). The Terrain dataset collection process is beyond the scope of this work; however, these data are made publicly available along with the code.

\subsection{Boids Coordinate dataset} 
\begin{figure}[tb]
\centering
\includegraphics[width=1\linewidth, keepaspectratio]{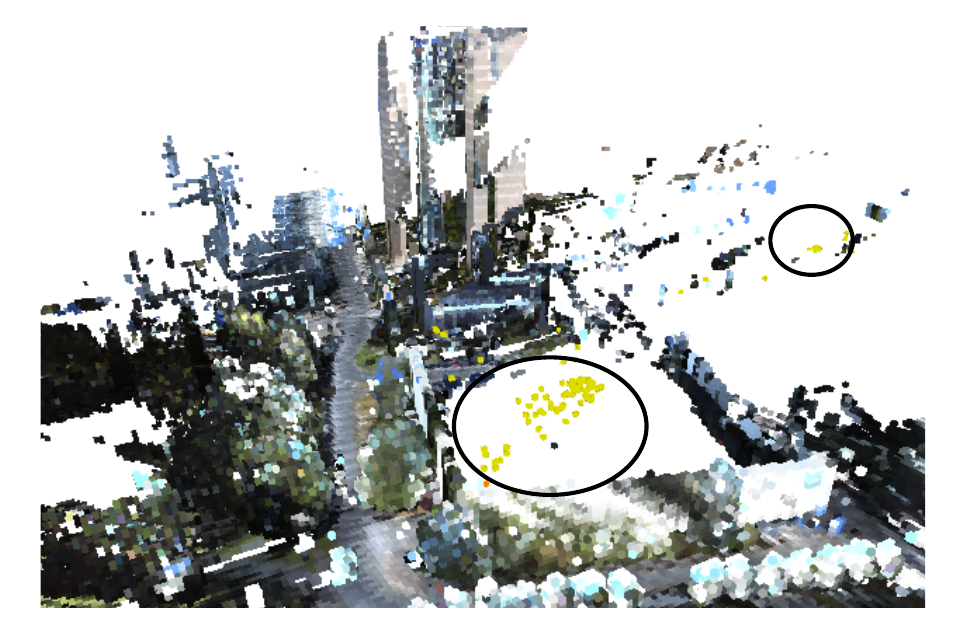}
\caption{A realistic 3D CubeletWorld featuring various terrain entities such as tall buildings (in gray in the background), streets running vertically with trees alongside, and various static entities. Clusters of yellow coloured boids (in circle) are dispersed throughout different regions of the CubeletWorld.}
\label{CWspreadOut}
\end{figure}

Boids Coordinate dataset is collected over approximately 10k samples by simulating the movement of $s = 30$ boids in the CubeletWorld. Each sample consists of a $P^{s \times 3}$ matrix representing the $(x, y, z)$ coordinates of all boids at a given time step. Figure \ref{CWspreadOut} illustrates groups of boids navigating through the environment.
Together, the Terrain and Boid Coordinate datasets create a dynamic simulation environment to train models to predict cubelet occupancy over time.

\section{Deep Architectures for CubeletWorld}

\begin{figure*}[h]
\centering
\includegraphics[width=.85\linewidth, keepaspectratio]{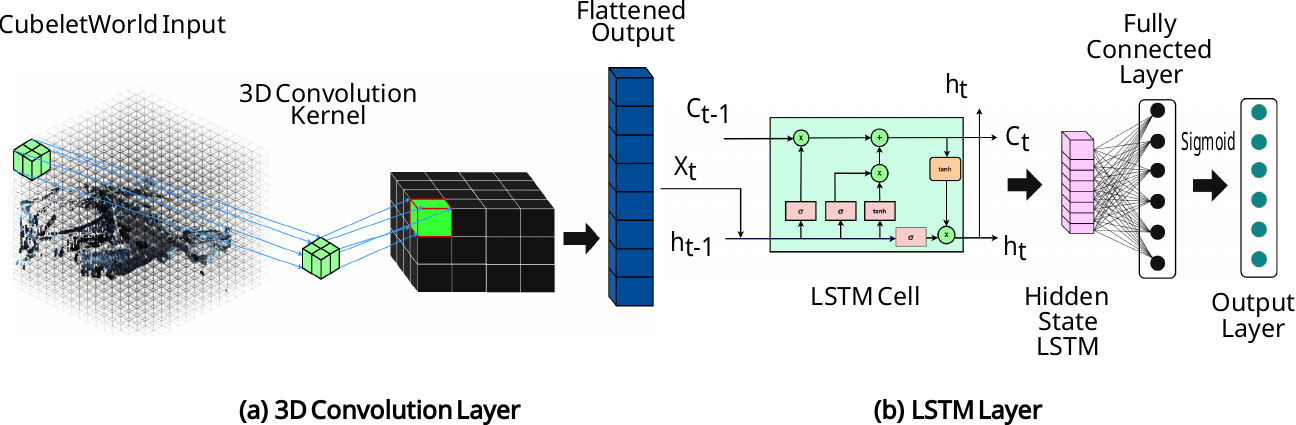}
\caption{Architecture of 3DCNN-LSTM model: The 3DCNN-LSTM model has two layers: 3D Convolution and LSTM layer. (a) A 3D convolution kernel convolves over the CubeletWorld sample, producing an output that is then flattened. (b) The flattened output then passes through the LSTM layer. The LSTM layer outputs the hidden state, which is then flattened and passed through a fully connected layer followed by a sigmoid function, which predicts the occupancy of each cubelet.}\label{cnnlstmmodel}
\end{figure*}

\subsection{CNN-LSTM Model} 
CNN-LSTM models have been shown to perform effectively on data involving spatial and temporal features, and thus they have found applications in various domains such as time series forecasting \cite{borovykh2017conditional}, medical imaging \cite{islam2020combined, shahzadi2018cnn}, and visual recognition \cite{donahue2015long, venugopalan2015sequence}.
\paragraph{Model Architecture}
The 3DCNN-LSTM architecture, as shown in Figure \ref{cnnlstmmodel} contains consists mainly of the following components:
\begin{itemize}[leftmargin=*]
    \item \textbf{3D Convolutional Layer (3DCNN):} This layer captures the spatial features of volumetric data, such as the CubeletWorld environment. It receives the cubelet representation of the environment as input. 
    \item \textbf{Long-Short Term Memory (LSTM):} LSTMs are good at capturing long-term dependencies in sequential data. We leverage this ability to capture the temporal relationship between states through time.
    \item \textbf{Fully Connected:} This layer simply combines the features learnt in previous layers and produces an output which is transformed via a sigmoid function to predict occupancy of each cubelet.
\end{itemize}

Figure \ref{cnnlstmmodel} describes our model pipeline. In order to predict for $n$ time steps in the future, we append the predicted labels to the input to predict the following time step. We continue this cycle recursively to get $n$ time step predictions. 

\subsection{Graph Neural Network Model}

The prediction of cubelet occupancy in CubeletWorld is accomplished using the Attention Temporal Graph Convolutional Network (A3T-GCN) \cite{bai2021a3t}. This model combines the strengths of graph convolutional networks (GCNs), gated recurrent units (GRUs), and an attention mechanism to capture the complex spatiotemporal dependencies inherent in the system.

\paragraph{Graph Construction}

\begin{figure}[tb]
    \centering
    \includegraphics[width=\linewidth, keepaspectratio]{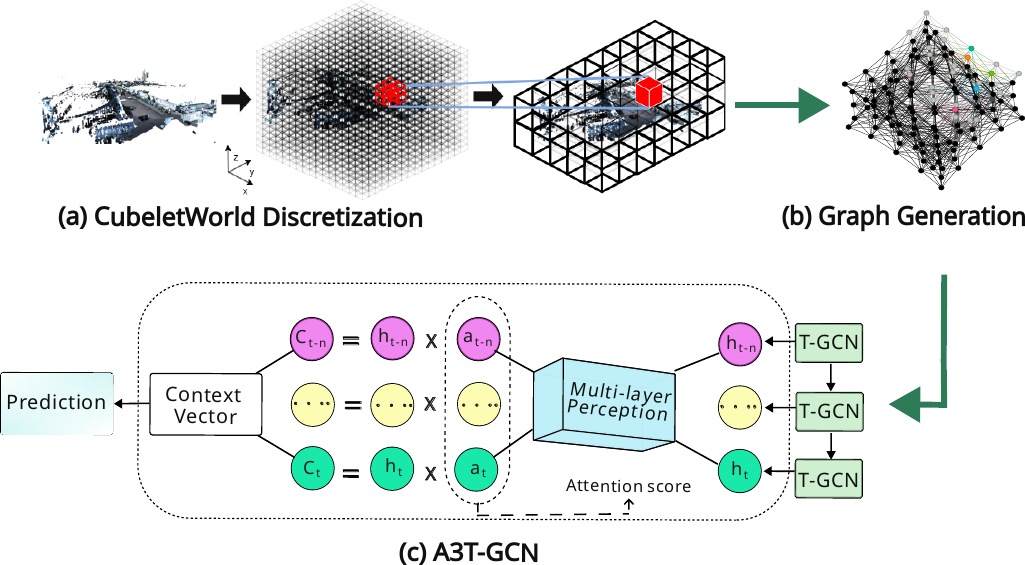}
    \caption{Architecture of A3T-GCN model: (a) The discretized CubeletWorld is first converted to a graph. (b) The generated graph is used as input to the A3T-GCN model. (c) The A3T-GCN model combines Temporal Graph Convolutional Network (T-GCN) \cite{zhao2019t} with an attention mechanism. Here, $h_t$ and $c_t$ denote hidden and cell states, while $a_t$ (and $a_{t-n}$) are attention scores over past hidden states, forming a context vector $C_t$ for predicting future cubelet occupancy.}

    \label{fig:gnn-diagram}
\end{figure}

CubeletWorld is represented as a dynamic graph, where each node corresponds to a cubelet, and edges represent spatial relationships such as adjacency or shared attributes. Depending on the size of the cubelets, we employ two different approaches for graph construction:

\begin{itemize}[leftmargin=*]
    \item \textbf{Full Graph:} For larger cubelet sizes (e.g., $4\times4\times4$, $5\times5\times5$, or greater), the entire spatial structure is treated as a full graph. To maintain computational efficiency, we remove nodes that are never occupied (i.e., those with zero values at all timestamps). We retain only the nodes that are occupied at least once during the sequence, along with their $k$-hop neighbors, even if those neighbors are never directly occupied. In our experiments in Section~\ref{experiments}, we set $k=1$, i.e., each occupied node is expanded to include its 1-hop neighbors in the full graph.

    \item \textbf{Subgraph Decomposition:} For smaller cubelet sizes (e.g., $3\times3\times3$ or smaller), processing the full graph becomes computationally prohibitive due to the sheer size of the graph (exceeding 1 million nodes and 10 million edges). In this case, the graph is partitioned into $k$-hop subgraphs, where each subgraph focuses on a central cubelet and its neighbouring nodes within $k$ hops (see Figure~\ref{fig:gnn-diagram} (b)). These subgraphs are processed independently and in parallel across multiple GPUs, ensuring scalability and efficiency. In our experiments in Section~\ref{experiments}, we use $k=2$ for multi-subgraph decomposition, i.e., each subgraph includes the central node and its 2-hop neighbors. Each subgraph is fed into an independent instance of the A3T-GCN, which performs localized spatial-temporal modeling. The predictions are obtained per subgraph without reassembling them into a single global graph. For evaluation, the results from all subgraphs are combined only at the metric level (e.g., averaging precision, recall, and F1-score across subgraphs).
\end{itemize}

This dual approach ensures that the graph construction process is adaptable to different cubelet sizes, enabling the model to effectively capture both spatial and temporal dependencies across CubeletWorld's varying resolutions.

\paragraph{Model Architecture}

We leverage the A3T-GCN architecture~\cite{bai2021a3t}, presented in Figure~\ref{fig:gnn-diagram}, as our core GNN, which comprises three primary components :

\begin{itemize}[leftmargin=*]
    \item \textbf{Graph Convolutional Layer (GCN):} This layer processes the graph structure to aggregate spatial information. By utilizing adjacency matrices and node features, the GCN learns the topological relationships between cubelets and their neighbours.
    \item \textbf{Gated Recurrent Unit (GRU):} Temporal dependencies are captured using GRUs, which process sequential data to learn dynamic changes in cubelet occupancy over time. GRUs are chosen for their simplicity and efficiency in handling time-series data.
    \item \textbf{Attention Mechanism:} An attention layer assigns weights to different temporal snapshots, emphasizing the most informative time steps. This mechanism improves the model’s ability to focus on relevant temporal features, enhancing prediction accuracy.
\end{itemize}

The combined architecture processes cubelet occupancy data through the GCN, refines temporal trends with the GRU, and dynamically adjusts temporal weights via the attention mechanism. Finally, the output layer predicts future cubelet occupancy states.

\section{Experiments}
\label{experiments}

We conduct occupancy prediction on the Cubelet Boids dataset. The CubeletWorld is discretized into cubelets at varying resolutions to evaluate the impact of granularity on model performance. For both datasets, we employ 3DCNN-LSTM and GNN-based models to predict the 10 time steps based on the previous 10 steps. To ensure robust evaluation, we perform a 5-fold cross-validation and report the average of the evaluation metrics across all folds.

\subsection{Data Processing}
We construct a 3D grid environment by mapping fixed entities in the Terrain dataset to CubeletWorld. Then, we simulate flocking behavior using boid dynamics to generate the Boids Coordinate dataset for the occupancy prediction task. This environment is configurable, allowing us to vary parameters, such as initial and maximum velocity of the boids, neighborhood radius, cohesion, and separation values of the environment, to produce diverse simulations. 
The size of the CubeletWorld for boids spans dimensions of $(827, 748, 173)$, with $s = 30$ boids navigating the space. The data is represented as a discretized input tensor $X \in R^{ t_1\times n_1\times n_2\times n_3}$, where $n_1$, $n_2$, and $n_3$ are the number of cubelets along each axis, and $t_1$ is the number of historical time steps. The target tensor is $Y \in R^{t_2 \times n_1\times n_2\times n_3}$, representing occupancy labels of each cubelet in the next $t_2$ time step. We chose $t_1=t_2=10$ in our experiments. Input $X$ is a binary matrix, where 1 denotes occupancy by at least one boid, and 0 otherwise. See Figure \ref{fig:cubeletdataset} for details. The models are trained to predict $Y$ based on $X$ as input. In total, each fold contains 7985 training and 1997 test samples.

\subsection{Evaluation Metrics}
We evaluate model performance using standard classification metrics: \textbf{Accuracy}, \textbf{Precision}, \textbf{Recall}, and \textbf{F1-score}. Accuracy is the ratio of correctly classified cubelets (both occupied and non-occupied) to the total number of cubelets. Precision is the ratio of correctly classified occupied cubelets to all cubelets predicted as occupied. Recall is the proportion of correctly predicted occupied cubelets from the total occupied cubelets. The F1 score is the harmonic mean of precision and recall.

\begin{table}[tb]
    \caption{Results of occupancy prediction on the CubeletWorld Boids dataset using the 3DCNN-LSTM and A3T-GCN model, with the number of future timesteps being 10. FG and MSG stand for Full Graph and Multi-Sub-Graphs, respectively. We denote "-" to indicate the infeasibility of experimentation due to scalability issues for the 3DCNN-LSTM model.}
    \label{tab:cnnboid}
    \centering
    \resizebox{\linewidth}{!}{
    \begin{tabular}{@{}llllllll@{}}
    \toprule
    Model          & Size of & Accuracy & Precision & Recall & F1-Score  \\
                   & each cubelet &     &           &        &           \\ 
    \hline
    CNN-LSTM       & (103, 93, 21) & 0.9989 & 0.9619 & 0.9645 & 0.9632  \\
    A3T-GCN (FG)        &               & 0.9616   & 0.7937  & 0.8338 & 0.8122 \\
    \hline
    CNN-LSTM       & (51, 53, 57) & 0.9969 & 0.8872 & 0.8482 & 0.8645 \\
    A3T-GCN (FG)        &              & 0.8902   & 0.6439    & 0.8319 & 0.6866 \\
    \hline
    CNN-LSTM       & (20, 20, 20) & 0.9996 & 0.7869 & 0.7714 & 0.7790 \\
    A3T-GCN (FG)       &              & 0.9727 & 0.6108 & 0.7296 & 0.6478 \\
    \hline
    CNN-LSTM       & (15, 15, 15) & 0.9999 & 0.3869 & 0.1885 & 0.1983 \\
    AT3-GCN (FG)       &              & 0.9791 & 0.5833 & 0.7113 & 0.6178 \\
    \hline
    CNN-LSTM       & (5, 5, 5)   &   -     &   -    &    -   &   - \\
    A3T-GCN (FG)       &             & 0.9972   & 0.5527    & 0.6358 & 0.5758 \\
    \hline
    CNN-LSTM       & (3, 3, 3)   &   -     &   -    &    -   &   - \\
    Multi A3T-GCNs
    (MSG)&             & 0.9611   & 0.6405    & 0.6647 & 0.5916 \\
    \bottomrule
    \end{tabular}
    }
\end{table}

\subsection{Results}
\label{result}

We report the result for the CubeletWorld Occupancy Prediction task in Table \ref{tab:cnnboid}. We perform experiments for different resolutions of cubelet size to observe their effect on the models' performance.

For the CNN-LSTM model, we observe that it strikes a good balance between recall and precision for all the cases except for the highest resolution, as shown in Table \ref{tab:cnnboid}.
However, the performance deteriorates as we decrease the size of the cubelet (or increase the resolution). We observe a steady fall in precision, recall, and F1 scores with increasing resolution, and this fall is quite significant in the highest-resolution case. 
Increasing the resolution makes the input \textit{sparser} as most cubelets lie unoccupied, and therefore, the model tends to predict zeros (or unoccupied) for the majority of cubelets, leading to a decrease in all the metrics except accuracy, which remains more or less the same in all the cases. Consider the first row of Table \ref{tab:cnnboid}, and the model has to make predictions on a total of only $9\times9\times9 = 729$ cubelets, while for the fourth row, there are a total of $56\times 50\times 12=33600$ cubelets. There is a 46-fold increase in the number of cubelets, and the fact that there are only 30 boids makes the cubelets sparsely occupied. 
Therefore, the model tends to predict zeros for most cases, leading to a huge decline in metrics. 
To add to the complexity, there might be clusters of boids spread across regions away from each other, as shown in Figure \ref{CWspreadOut}. 
Another reason for the decline in the performance of the 3DCNN-LSTM model could be attributed to \textit{error propagation} due to the sequential nature of prediction. Since we append the prediction at time step $t$ to the input in order to make a prediction for time step $t+1$, any wrong predictions can lead to worse predictions at later time steps. The combined effect of the issues highlighted above is intensified in the higher-resolution cases.
We also decided to conduct experiments for even smaller resolutions; however, given the huge number of cubelets, training and evaluation become computationally prohibitive for the 3DCNN-LSTM model, indicating that the model does not scale well with an increased number of cubelets. The number of cubelets for the last row in Table \ref{tab:cnnboid} is $276\times 250\times 58=4e6$ and therefore computationally infeasible to train the 3DCNN-LSTM model in a comprehensible time. This highlights the challenges when making fine-grained predictions in the CubeletWorld, namely, scalable training of models while ensuring that models can handle sparsity in the environment.

The experimental results in Table \ref{tab:cnnboid} for A3T-GCN show a trend similar to the 3DCNN-LSTM model. We observe a steady decline in the evaluation metrics with increasing resolution of the 3D grid world. 
We observe that recall is higher than precision for all the cases, indicating that the model is not so cautious in labelling a cubelet positive (or occupied). The model prefers to label a cubelet occupied at the cost of wrongly labeling some cubelets. However, this is not enough to affect accuracy, as it remains high in all the cases; this is due to the majority of cubelets being unoccupied, and nevertheless, it still tries to predict zeros (or unoccupied) for most cubelets. 
Compared to 3DCNN-LSTM, we do not observe a huge decline in the case of cubelet size 15. For the lower resolution, the 3DCNN-LSTM model seems to be a better choice. However, it is not scalable. For smaller cubelet sizes (or higher resolution) (e.g., 3x3x3), we performed a subgraph decomposition to deal with the issue of scalability and observed that the subgraph model actually performs better or equally well compared to some lower-resolution cases. The model strikes a balance between being neither too conservative nor too liberal in predicting the positives. However, the F1-score indicates that while this model is good at striking a balance, it still struggles to identify positive classes. 
This improvement arises because each subgraph is processed independently by an A3T-GCN, which constrains message passing to a local neighborhood and avoids the prohibitive cost of modeling the entire million-node graph. By focusing on localized interactions, the model reduces noise from distant, irrelevant nodes and keeps the receptive field compact. This not only makes training feasible at fine resolutions but also contributes to the observed performance stability in high-resolution settings. Overall, these results highlight that subgraph decomposition is essential for making CubeletWorld predictions tractable at high resolutions, enabling the A3T-GCN to operate effectively where full-graph methods become infeasible.

\section{Discussions and Limitations}\label{discussionandlimitations}

In Section \ref{result}, we highlighted the challenge of model performance degradation with increasing spatial granularity and the difficulty of scaling predictions to a larger number of cubelets. As the size of individual cubelets decreases, the state representations become sparser, and the number of cubelets the model must make predictions for per sample increases. We elaborate on these results and their broader implications below: 
\begin{itemize}[leftmargin=*]
    \item We observed consistently high accuracy across all cases, largely attributable to the sparse nature of the state representations. Notably, the CNN-LSTM model outperformed the A3T-GCN in lower-resolution tasks, suggesting its suitability for domains such as drone trajectory planning or socio-economic modeling, where one might predict migration patterns across urban neighborhoods. 
    \item However, this model also exhibits scalability issues, with performance diminishing as granularity increases. This underscores the importance of balancing resolution and scalability, an essential consideration that should be tailored to the application domain. For example, lower-resolution predictions may suffice for summarizing broader regions in low-risk scenarios such as urban socio-economic assessments (e.g., estimating income levels across city neighborhoods). Conversely, in critical domains like autonomous vehicle routing or disaster response planning, where high-resolution predictions are necessary, the performance decline with increased granularity poses a significant challenge.
    \item Given the strong performance of the subgraph model and its inherent ability to address scaling challenges, one promising direction is to train local prediction models specific to different regions of the 3D discretized grid. This could involve training multiple region-specific models or leveraging transfer learning to enable a single model to generalize across regions with minimal fine-tuning. Alternatively, hierarchical models can be used to process and predict at multiple levels of abstraction, effectively handling both resolution and scalability by operating on smaller segments of the environment.
    \item Treating subgraphs as independent inputs to separate A3T-GCN instances avoids the need to reconstruct a global graph and reduces error propagation across regions. This design enables straightforward parallelization across GPUs, making the approach both computationally tractable and robust when scaling to millions of cubelets. In particular, subgraph decomposition substantially reduces the computational burden of huge graphs (e.g., $3\times3\times3$ cubelets with over one million nodes), allowing training and inference to remain feasible.
    \item As mentioned in Section \ref{sec1}, the cubelet-centric approach in a 3D grid world offers potential advantages for preserving data privacy. However, this benefit is contingent upon the chosen level of granularity. Even when data are abstracted into cubelets, high-resolution representations, or sensitive data types can still present reidentification risks. Robust anonymization techniques and privacy-preserving methods must be employed, particularly at finer resolutions, to mitigate potential deanonymization threats.
\end{itemize}

\section{Related Work}\label{relatedworks}
\subsection{Semantic Occupancy Prediction} 
3D Semantic Occupancy Prediction refers to the task of determining whether a voxel is occupied and assigning it a semantic label. This research area is particularly relevant for vision-based autonomous driving, using a single agent from multi-view imagery or LiDAR streams \cite{zuo2023pointocc, li2022bevformer, huang2023tri}. Semantic scene understanding entails reconstructing the entire spatial scene, recognizing relationships between entities, and semantically labeling each object. 
Numerous datasets have been proposed for this task \cite{Caesar2019nuScenesAM, Behley2019SemanticKITTIAD, Liao2021KITTI360AN}. Some rely solely on images \cite{Geiger2012AreWR}, while others incorporate richer modalities such as LiDAR, radar, and multi-view imagery \cite{Caesar2019nuScenesAM, Behley2019SemanticKITTIAD, Liao2021KITTI360AN}. Li \textit{et al.} \cite{li2022bevformer} and Huang \textit{et al.} \cite{huang2023tri} both use transformer models with a 2DCNN encoder to combine features from multi-view images and LiDAR (only the latter). Unlike these agent-centric approaches, our focus lies in modeling the CubeletWorld as a whole. Rather than reconstructing the scene from an agent's perspective and predicting voxel-wise semantics, our model receives the entire CubeletWorld as input for occupancy prediction.

Additionally, several datasets \cite{Pellegrini2009YoullNW, Lerner2007CrowdsBE, robicquet2016learning} have been introduced for trajectory prediction tasks. Alahi \textit{et al.} \cite{alahi2016social}, for instance, employed an LSTM-based model to learn human movement patterns and predict future trajectories, demonstrating the value of temporal modeling in dynamic environments. 
4D Occupancy Prediction is another closely related task where the goal is to forecast the future evolution of the surroundings: moving agents as well as static entities in the scene. This area of research still significantly differs from our work since it is agent-centric. Our work aims to complement these vision-based tasks by incorporating heterogeneous data into a global 3D representation.

\subsection{CNN-LSTM Models}
The combination of CNN and LSTM networks forms a powerful architecture for capturing both spatial and temporal dependencies in sequential data.  This hybrid architecture has been applied successfully in diverse domains, including visual recognition and description \cite{donahue2015long, venugopalan2015sequence}, time series forecasting \cite{borovykh2017conditional}, rumor detection \cite{azri2021calling}, predictive maintenance \cite{zhou2021automatic}, and medical image analysis \cite{islam2020combined, shahzadi2018cnn}.  

CNN-LSTM architectures have also shown success in volumetric classification tasks. For example, Shahzadi \textit{et al.} \cite{shahzadi2018cnn} applied a CNN-LSTM model to classify 3D brain tumor MRI volumes into high and low grade gliomas, using VGG-16 for spatial feature extraction.
CNN-LSTM architectures offer strong capabilities in modeling spatio-temporal patterns, making them a compelling choice for occupancy predictive tasks in the CubeletWorld setting.

\subsection{Graph Neural Networks}
Graph Neural Networks (GNNs) are well-suited for modeling structured data in sensor networks, where entities and their relationships can be naturally represented as graphs. GNNs have demonstrated success in tasks such as occupancy prediction, traffic flow monitoring, and environmental sensing.
They typically model sensors or regions as nodes, with edges capturing spatial or functional relationships. Wu \textit{et al.} \cite{Wu2020ComprehensiveSurvey} provided a taxonomy of GNN architectures, including convolutional and recurrent variants, which effectively model inter-node interactions. Li \textit{et al.} \cite{Li2021TrajectoryPrediction} applied GNNs to trajectory prediction, showing their capacity to capture both spatial and temporal dynamics in sensor data.

GNNs have also been applied to multiple sensors and dynamic scenarios. For example, Yu \textit{et al.} \cite{Yu2017STGCN} developed a spatio-temporal GNN for traffic prediction, integrating spatial and temporal signals to enhance forecasting accuracy. Zhang \textit{et al.} \cite{Zhang2020ParkingPrediction} extended this to dynamic IoT environments, demonstrating the adaptability of GNNs to evolving inter-sensor relationships.

In the context of autonomous systems, GNNs have shown utility in modeling complex interactions. Casas \textit{et al.} \cite{Casas2018VehicleInteraction} used GNNs to model vehicle interactions, improving trajectory prediction accuracy. Similarly, Mohamed \textit{et al.} \cite{Mohamed2020PedestrianMovement} proposed a graph-based model for pedestrian movement, integrating spatial and temporal features for robust prediction. 
These examples illustrate the strong potential of GNNs for the CubeletWorld framework, where the objective is to predict cubelet occupancy based on historical states and the configuration of neighboring cubelets. Their ability to model complex spatial-temporal dependencies makes GNNs a promising direction for scalable and structured prediction in 3D environments.

Building on this line of research, Zhao \textit{et al.} \cite{zhao2019t} introduced the Temporal Graph Convolutional Network (T-GCN) to jointly model spatial dependencies with graph convolutions and temporal dynamics with recurrent units. Bai \textit{et al.} \cite{bai2021a3t} further extended this idea with an attention mechanism (A3T-GCN). Our work adapts this architecture to CubeletWorld by applying it independently to decomposed $k$-hop subgraphs, a strategy that mirrors neighborhood-based designs in traffic and IoT forecasting while addressing the scalability challenges of 3D environments.

\section{Conclusion and Future Work}\label{conclusionandFW}
We proposed a novel method for representing and storing information in volumetric spaces by discretizing them into a 3D grid, the \textit{CubeletWorld}. This representation has broad applicability across domains such as route planning, environmental monitoring (e.g., identifying green cover or pollution levels), socioeconomic studies, and last but not least, pandemic spread. 
The \textit{CubeletWorld} enables storage of multi-modal data within a shared spatial framework, supports multiple levels of granularity, while preserving privacy. We propose to shift from agent-centric approaches to a 3D global representation, thereby complementing other vision-based tasks. To demonstrate this approach, we developed \textit{Cubelet Boids dataset}, focusing on occupancy prediction. Our core models, CNN-LSTM and A3T-GCN, highlighted key challenges that accompanies fine-grained prediction. Higher resolution leads to sparser representations and an exponential rise in the number of cubelets, thereby degrading model performance and complicating training at scale.
In the future, we aim to address these challenges through two key directions. First, we plan to develop region-specific models that can be transferred and adapted to other areas of the environment. Second, we propose exploring hierarchical models capable of processing data at multiple granularities. These strategies hold promise for more scalable and accurate predictive models in complex 3D environments.
Combining heterogeneous data stored in the cubelet could also be promising for better and interpretable predictions and analysis. However, one needs to be cautious when doing fine-grained analysis at individual apartments, indoor space, as compared to coarser analysis that aggregates data over larger cubelets covering entire city blocks or buildings. 
By addressing these current challenges, this paradigm has the potential to advance and complement predictive modeling and decision-making in complex 3D environments, ultimately contributing to better urban and environmental systems.

{
    \small
    \bibliographystyle{ieeenat_fullname}
    \bibliography{sn-bibliography}
}

\end{document}